# Robust Audio Event Recognition with 1-Max Pooling Convolutional Neural Networks

Huy Phan*†, Lars Hertel*, Marco Maass*, *and* Alfred Mertins*

*Institute for Signal Processing, University of Lübeck
†Graduate School for Computing in Medicine and Life Sciences, University of Lübeck
{phan,hertel,maass,mertins}@isip.uni-luebeck.de

## Abstract

We present in this paper a simple, yet efficient convolutional neural network (CNN) architecture for robust audio event recognition. Opposing to deep CNN architectures with multiple convolutional and pooling layers topped up with multiple fully connected layers, the proposed network consists of only three layers: convolutional, pooling, and softmax layer. Two further features distinguish it from the deep architectures that have been proposed for the task: varying-size convolutional filters at the convolutional layer and 1-max pooling scheme at the pooling layer. In intuition, the network tends to select the most discriminative features from the whole audio signals for recognition. Our proposed CNN not only shows state-of-the-art performance on the standard task of robust audio event recognition but also outperforms other deep architectures up to 4.5% in terms of recognition accuracy, which is equivalent to 76.3% relative error reduction.

## 1. Introduction

The success of deep architectures in many applications is explained by their ability to discover multiple levels of features from data. Inspired by this, many deep neural networks have recently been proposed for audio event recognition. In [1, 2], deep neural networks (DNNs) are first initialized using unsupervised training with deep belief networks (DBNs) [3] and then trained by the standard backpropagation. In order to deal with event overlap, DNNs with multi-label classification schemes have also been proposed [4]. Recently, various deep CNN architectures with multiple convolutional and pooling layers for hierarchical feature extraction have also been employed [5, 6, 7, 8]. Although these deep networks showed promising performance, especially under difficult conditions such as under interference [1, 6] and event overlapping [4], they come with a significant shortcoming. These deep architectures require equal-size inputs while the nature of audio events exhibits high intra- and inter-class temporal durations. To go around this issue, the signals were decomposed into equal segments and the models were then trained on these local features. In turn, the evaluation also took place on local features followed by some voting schemes, e.g. majority voting [1, 7, 8] and probability voting [7], to obtain a global classification label. Although this adaptation helps to facilitate the training and testing of the models, it is incapable of capturing the *shift-invariance* property [9] that the cochlea and auditory nerve in the auditory system have [10]. This is really undesirable since a particular feature could be replicated at *any* time in the signal instead of its local segments.

We present a convolutional neural network architecture for robust audio event recognition that is able to address these issues. Our architecture is much simpler and more "shallow". It consists of three layers: convolutional, pooling, and softmax layer. The convolution layer coupled with the pooling layer are responsible for feature extraction and the final softmax layer is in charge of classification. Our proposed architecture is different from the deep ones that have been used for the task in many aspects. Foremost, it takes the whole signals of audio events as input instead of their small fractions. Second, we do not fix the size of the convolutional filters at the convolutional layer as in conventional CNNs but allow multiple filters with different sizes to be learned simultaneously. Consequently, we are able to capture features at multiple resolutions of audio signals. Third, we do not pursue subsampling at the pooling layer but *1-max pooling* scheme. As a result, with the feature map induced by convolving one of the filters on an input signal, we only select the most prominent feature. The prominent features produced by all filters are finally concatenated and presented to the final softmax layer for classification. Furthermore, owing to the 1-max pooling, the inputs to the network can be of any arbitrary size. That is, we can naturally deal with the intra- and inter-class temporal variation of audio events. Lastly, each convolutional filter can be thought of playing the role of a cochlear filter which spikes on a specific feature of the signal [11, 10]. In addition, the feature is allowed to happen at any time in the signal, i.e. it is shift-invariant.

## 2. The proposed approach

In this section we will present the spectrogram image features that are used to represent audio signals. Afterwards, our proposed CNN architecture will be described. The spectrogram images are used as inputs for the network.

### 2.1. Spectrogram image features (SIF)

Given an audio signal, it is decomposed into overlapping segments from which a spectrogram is generated by short-time Fourier transform. The short-time spectral column representing a length-$L$ segment $s_t(n)$ at the time index $t$ is given by

$$\mathbf{S}(f,t) = \left| \sum_{n=0}^{L-1} s_t(n)\phi(n) e^{\frac{-j2\pi nf}{L}} \right|, \quad (1)$$

where $f = 0, \ldots, (\frac{L}{2} - 1)$ and $\phi(n)$ denotes a $L$-point Hamming window. The spectrogram is then down-sampled in frequency to keep a $F$-bin frequency resolution by averaging over a window of length $W = \lfloor L/2F \rfloor$.

A de-noising step is finally performed by subtracting the minimum value from each spectral vector over time:

$$\mathbf{S}_{dn}(f,t) = \mathbf{S}(f,t) - \min_t(\mathbf{S}(f,t)), \quad (2)$$

for $f = 0, \ldots, (F-1)$. The short-time energy $e(t)$ can also be appended to the spectrogram image as an augmented feature:

$$e(t) = \sum_{f=0}^{F-1} \mathbf{S}_{dn}(f,t). \quad (3)$$

Our proposed SIF features are similar to those in [1]. However, instead of classifying on equal spectro-temporal patches of the images, our classification is efficiently performed on the whole varying-size spectrogram images.

### 2.2. 1-Max Pooling CNN

The proposed network consists of three layers, including convolutional, pooling, and softmax layer as illustrated in Figure 1.

#### 2.2.1. Convolutional layer

We aim to use the convolutional layer to extract discriminative features within the whole signals that are useful for the classification task at hand. Suppose that a spectrogram image presented to the network is given in the form of a matrix $\mathbf{S} \in \mathbb{R}^{F \times T}$ where $F$ and $T$ denote the number of frequency bins and the number of audio segments, respectively. We then perform convolution on it via linear filters. For simplicity, we only consider convolution in time direction, i.e. we fix the height of the filter to be equal to the number of frequency bins $F$ and vary the width of the filter to cover different number of adjacent audio segments.

Let us denote a filter by the weight vector $\mathbf{w} \in \mathbb{R}^{F \times w}$ with the width $w$. Therefore, the filter contains $F \times w$ parameters that need to be learned. We further denote the adjacent spectral columns (e.g. audio segments) from $i$ to $j$ by $\mathbf{S}[i:j]$. The convolution operation $*$ between $\mathbf{S}$ and $\mathbf{w}$ results in the output vector $\mathbf{O} = (o_1, \ldots, o_{T-w+1})$ where

$$o_i = (\mathbf{S} * \mathbf{w})_i = \sum_{k,l} (\mathbf{S}[i : i+w-1] \odot \mathbf{w})_{k,l}. \quad (4)$$

Here, $\odot$ denotes the element-wise multiplication. We then apply an activation function $h$ to each $o_i$ to induce the feature map $\mathbf{A} = (a_1, \ldots, a_{T-w+1})$ for this filter:

$$a_i = h(o_i + b), \quad (5)$$

where $b \in \mathbb{R}$ is a bias term. Among the common activation fuctions, we chose *Rectified Linear Units* (ReLU) [12] due to their computational efficiency:

$$h(x) = \max(0, x). \quad (6)$$

To allow the network to extract complementary features and enrich the representation, we learn $P$ different filters simultaneously. Furthermore, the use of multiple resolution levels has been shown important for the task [5] as the time duration that yields salient features may vary depending on the event categories. In order to account for this, we learn $Q$ different sets of $P$ filters, each of which has different width $w$ to form totally $Q \times P$ filters.

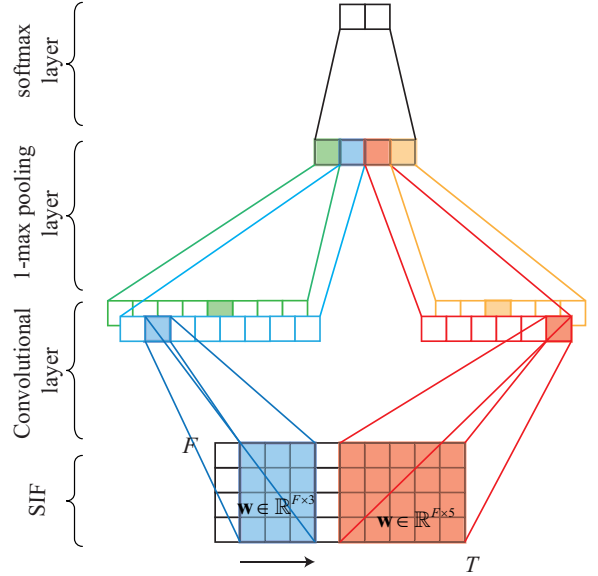

Figure 1: Illustration of 1-max pooling CNN architecture. The network consists of two filter sets with two different widths $w = \{3, 5\}$ at the convolutional layer. There are two individual filters on each filter set.

#### 2.2.2. 1-max pooling layer

The feature maps produced by the convolution layer are forwarded to the pooling layer. We employ 1-max pooling function [13] on a feature map to reduce it to a single most dominant feature. Pooling on $Q \times P$ feature maps results in $Q \times P$ features that will be joined to form a feature vector inputted to the final softmax layer. This pooling strategy offers a unique advantage. That is, although the dimensionality of the feature maps varies depending on the length of audio events and the width of the filters, the pooled feature vectors have the same size of $P \times Q$. The same strategy has recently been proved useful in different tasks of natural language processing owing to its ability to cope with varying-size input texts, such as sentences [14, 15]. Coupled with the 1-max pooling function, each filter in the convolutional layer is optimized to detect a specific feature that is allowed to occur at any time in a signal.

#### 2.2.3. Softmax layer

The fixed-size feature vector after the pooling layer is subsequently presented to the standard softmax layer to compute the predicted probability over the class labels. The network is trained by minimizing the cross-entropy error. This is equivalent to minimizing the KL-divergence between the prediction distribution $\hat{y}$ and the target distribution $y$. With the binary one-hot coding scheme and the network parameter $\theta$, the error for $N$ training samples is given by:

$$E(\theta) = -\frac{1}{N} \sum_{i=1}^{N} y_i \log(\hat{y}_i(\theta)) + \frac{\lambda}{2} ||\theta||^2. \quad (7)$$

The hyper-parameter $\lambda$ governs the trade-off between the error term and the $\ell_2$-norm regularization term. For regularization purposes, we also employ dropouts [16] at this layer by randomly setting values in the weight vector to zero with a predefined probability. The optimization is performed using the *Adam* gradient descent algorithm [17].

# 3. Experiments

## 3.1. Databases

We set up the standard experiment of the robust audio event recognition task similar to current state-of-the-art works [18, 1, 6] so that the results are comparable.

**Audio event database.** We targeted 50 sound event categories[1] from the Real Word Computing Partnership (RWCP) Sound Scene Database in Real Acoustic Environments [19]. For each category, we randomly selected 80 sound instances which were divided into 50 instances for training and 30 instances for testing. Out of 50 training instances, we left out 10 instances for validation, and other 40 instances were used to tune the networks. It turns out that there are totally 2000, 500, and 1500 event instances for training, validation, and testing purpose, respectively.

**Noise database.** As in [18, 1, 6], we chose four different environmental noises from NOISEX-92 database [20], including "Destroyer Control Room", "Speech Bable", "Factory Floor 1", and "Jet Cockpit 1". Beside clean signals, we also created noise-corrupted signals by randomly choosing one of four noise signals to add to the clean signals at random starting points. The noise signals were added with different level of 20, 10, and 0 dB signal-to-noise ratio (SNR). We evaluate both mismatched condition (tranining with only clean event instances) and multi-condition (training with both clean and noise-corrupted event instances).

## 3.2. Parameters

Audio signals sampled at 16 kHz sampling frequency were divided into 100 ms frames with a hop of 10 ms. Each frame was analyzed with 2048-point FFT to obtain a spectral column which is then down-sampled as described in Section 2.1 to keep $F = 52$ frequency bins. Although the SIFs can be of arbitrary sizes, we zero-padded them column-wise in time direction to ease the implementation.

The proposed CNN architecture involves different hyper-parameters which are specified in Table 1. Although the hyper-parameters were set to very common values, parameter search can be done to further enhance the performance. The networks were trained using the training set for 1000 epochs (mismatched condition) and 500 epochs (multi-condition) with a minibatch size of 100. During training the networks that maximize the classification accuracy on the validation set will be retained.

## 3.3. Classification systems

We trained four different networks using our proposed architecture:
- **1MaxCNN**: our proposed SIF and 1-max pooling CNN (mismatched condition).
- **1MaxCNN-E**: our proposed energy-augmented SIF and 1-max pooling CNN (mismatched condition).
- **1MaxCNN-MC**: our proposed SIF and 1-max pooling CNN (multi-condition).
- **1MaxCNN-E-MC**: our proposed energy-augmented SIF and 1-max pooling CNN (multi-condition).

We compare the classification accuracy against other systems [18, 1, 6] with the standard experimental setup. They include
- MFCC-HMM [18]: Mel Frequency Cepstral Coefficients (MFCC) with a Hidden Markov Models (HMM) backend.

---
[1]The specific event categories are based on unofficial communication with Jonathan W. Dennis, the author of [18].

Table 1: Hyper-parameters of the proposed CNN networks.

| Hyper-parameter | Value |
| --- | --- |
| Filter sizes | $\{1, 3, \ldots, 25\}$ |
| Number of filter $P$ for each size | 100 |
| Learning rate for the Adam optimizer | 0.0001 |
| Dropout rate | 0.5 |
| Regularization parameter $\lambda$ | 0.0001 |

- MFCC-SVM [18]: MFCC with a Support Vector Machine (SVM) backend.
- ETSI-AFE [18]: above MFCC-SVM that is further evaluated with an ETSI Advanced Front End toolkit enhancement [21].
- MPEG-7 [18]: a set of 57 low-level features coupled with Principle Component Analysis (PCA) feature selection and a HMM classifier.
- Gabor [18]: Gabor features followed by single-layer perceptron feature selection and HMM classification.
- GTCC [18]: Gammatone cepstral coefficients features with a HMM backend.
- MP+MFCC [18]: MFCCs and Gabor features from top five Gabor bases found by the matching pursuit (MP) algorithm [22] backed with a HMM classifier.
- Dennis SIF [18]: a similar SIF and a SVM classifier.
- SIF-DNN [1]: a similar SIF and DNN classification (mismatched condition).
- SIF-DNN-MC [1]: a similar SIF and DNN classification (multi-condition).
- SIF-CNN [6]: a similar SIF and deep CNN classification.
- SIF-IS-CNN [6]: an enhanced SIF by smoothing and deep CNN classification.
- SIF-IS-DNN [6]: an enhanced SIF by smoothing and DNN classification.
- MelFb-CNN [6]: an enhanced SIF features with Mel-filterbank analysis and deep CNN classification.

## 3.4. Experimental results

### 3.4.1. Performance as a function of the filter width

We show in Figure 2 the performance of our systems in terms of classification accuracy as a function of the filter width $w$ in different noise conditions. When $w$ varies from small to large values, the features learned by the networks are expected to change from detail to higher abstracted ones. As can be seen, in most of the cases the accuracies grow with the increase of $w$.

For the **1MaxCNN** system with mismatched condition, although it shows good robustness in low to mid-range noise conditions, it is less robust in harsh noise condition of 0 dB. In addition, when augmented with the short-time energy feature, the system **1MaxCNN-E** exhibits strong sensitivity in noise conditions. However, when being trained with multi-condition data, both **1MaxCNN-MC** and **1MaxCNN-E-MC** expose remarkably strong robustness to all noise conditions. The reason is that presenting the networks with mutli-condition data is not only about data augmentation but also enforces them to learn noise-robust filters.

### 3.4.2. Performance comparison

The comparison on classification accuracy of our systems and the competitive systems is given in Table 2. Note that although our systems with a single filter size trained with multi-condition data can easily outperform the best competitor, we use the systems with multiple filter widths in $\{1, 3, \ldots, 25\}$ (equivalent to $\{100, 120, \ldots, 340\}$ ms respectively) for comparison here. It is partly because of the clarity's sake and partly because these systems are able to capture features on multiple resolutions and offer even better performance.

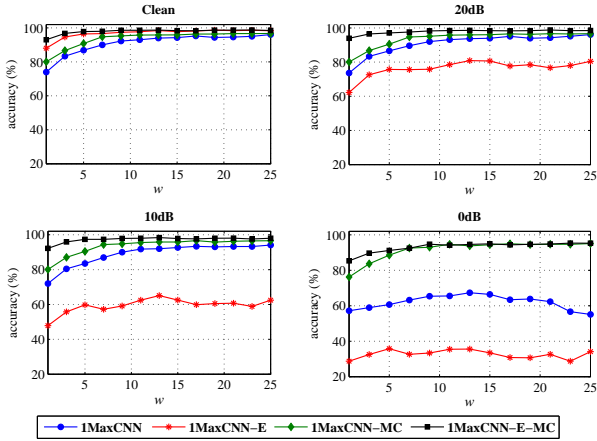

Figure 2: Classification accuracy as a function of the filter width $w$ for different noise conditions.

It can be seen that our system **1MaxCNN** performs significantly better than all deep-architecture opponents on clean, 20 dB, and 10 dB conditions although it is incomparable with the low-level feature systems (e.g. Gabor, GTCC) on the clean conditions and less robust than some deep architectures (e.g. SIF-CNN, SIF-DNN) in worst noise condition of 0 dB. Again, when augmented with short-time energy features, the system **1MaxCNN-E** exhibits its sensitivity in noise conditions although 1.1% absolute improvement can be seen in the noise-free condition.

On the other hand, our multi-condition trained systems **1MaxCNN-MC** and **1MaxCNN-E-MC** show superior performance compared to all deep-architecture opponents in all testing conditions, especially in the hardest one of 0 dB. Compared to the best deep-architecture competitor (i.e. SIF-IS-CNN), **1MaxCNN-MC** shows absolute gains of 1.1%, 1.0%, 2%, and 12.2% on noise-free, 20 dB, 10 dB, and 0 dB conditions, respectively. Those corresponding improvements obtained by **1MaxCNN-E-MC** are even better with 1.8%, 1.7%, 2.7%, and 12.0%. These lead to average absolute accuracy gains of 4.1% and 4.5% which are equivalent to relative error reduction rates of 69.5% and 76.3% for **1MaxCNN-MC** and **1MaxCNN-E-MC**, respectively. Given the fact that multi-condition training was reported to result in little benefit on the task (for example, SIF-DNN-MC compared to SIF-DNN [1]), the performance of our multi-conditioned systems are quite impressive.

### 3.5. Discussion

Our proposed 1-max pooling CNN shows very promising performance even though we conservatively set the hyper-parameters to very common values. Since there are many hyper-parameters (e.g. the activation function, the filter width, the number of filters, the learning rate, the dropout rate, the regularization term $\lambda$), the chance to find a better set of values for them via parameter tuning is actually large. Furthermore, it is also worth further analyzing the sensitivity of the networks to these hyper-parameter values.

Table 2: Classification accuracy (%) comparison (results of the competitive systems courtesy of [18, 1, 6]).

| System | clean | 20dB | 10dB | 0dB | mean |
|---|---|---|---|---|---|
| MFCC-HMM | 99.4 | 71.9 | 42.3 | 15.7 | 57.4 |
| MFCC-SVM | 98.5 | 28.1 | 7.0 | 2.7 | 34.1 |
| ETSI-AFE | 99.1 | 89.4 | 71.7 | 35.4 | 73.9 |
| MPEG-7 | 97.9 | 25.4 | 8.5 | 2.8 | 33.6 |
| Gabor | **99.8** | 41.9 | 10.8 | 3.5 | 39.0 |
| GTCC | 99.5 | 46.6 | 13.4 | 3.8 | 40.8 |
| MP+MFCC | 99.4 | 78.4 | 45.4 | 10.5 | 58.4 |
| Dennis SIF | 91.1 | 91.1 | 90.7 | 80.0 | 88.5 |
| SIF-DNN | 96.0 | 94.4 | 93.5 | 85.1 | 92.3 |
| SIF-DNN-MC | 94.7 | 95.8 | 92.1 | 87.7 | 92.6 |
| SIF-CNN | 97.3 | 97.4 | 95.7 | 83.1 | 93.4 |
| SIF-IS-CNN | 97.3 | 97.3 | 96.2 | 85.5 | 94.1 |
| SIF-IS-DNN | 86.7 | 86.4 | 85.3 | 73.5 | 83.0 |
| MelFb-CNN | 97.7 | 97.5 | 94.7 | 70.3 | 90.0 |
| **1MaxCNN** | 98.0 | **98.1** | **97.3** | 75.5 | 92.2 |
| **1MaxCNN-MC** | 98.4 | **98.3** | **98.2** | 97.7 | **98.2** |
| **1MaxCNN-E** | 99.1 | 88.5 | 74.9 | 50.3 | 78.2 |
| **1MaxCNN-E-MC** | 99.1 | **99.0** | **98.9** | 97.5 | **98.6** |

On the other hand, for simplicity we fixed the height of the filters equal to the number of frequency bins and only varied the width of the filters in time. And by this, we only conducted convolution in time direction. One possible improvement is to additionally allow convolution in frequency dimension, for example in different frequency subbands. However, the convolution should respect the order of the frequencies since it simply matters for audio signals. Lastly, it is also interesting to visualize the filters to see what the networks actually learn.

## 4. Conclusions

We presented a CNN network architecture that is efficient for robust audio event recognition. Compared to deep CNNs, our proposed architecture is relatively simple and more "shallow". Intuitively, with each convolutional filter coupled with 1-max pooling scheme, our CNNs based on the proposed architecture tend to extract the most discriminative and shift-invariant features from the audio signals for recognition. In addition, we can naturally deal with the temporal variations of audio events, thanks to the 1-max pooling scheme. In an evaluation on the standard task of robust audio event recognition, we obtain a relative error reduction of 76.3% compared to the reported results from the best deep CNN opponent.

## 5. Acknowledgements

This work was supported by the Graduate School for Computing in Medicine and Life Sciences funded by Germany's Excellence Initiative [DFG GSC 235/1].